\documentclass{article}

\usepackage{arxiv}

\usepackage[utf8]{inputenc} 
\usepackage[T1]{fontenc}    
\usepackage[colorlinks=true, citecolor=blue, linkcolor=blue, urlcolor=blue]{hyperref}
\usepackage{url}            
\usepackage{booktabs}       
\usepackage{amsfonts}       
\usepackage{nicefrac}       
\usepackage{amsmath}
\usepackage{microtype}      
\usepackage{cleveref}       
\usepackage{lipsum}
\usepackage{array}
\usepackage{graphicx} 
\usepackage{subcaption}
\usepackage{doi}
\usepackage{xcolor}
\usepackage{authblk}

\newbox{\orcid}
\sbox{\orcid}{\includegraphics[scale=0.06]{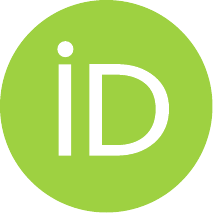}}

\title{Involvement drives complexity of language in online debates}

\author[1]{\href{https://orcid.org/0000-0000-0000-0000}{\usebox{\orcid}}\hspace{1mm}Eleonora Amadori}
\author[2]{\href{https://orcid.org/0009-0001-6727-3316}{\usebox{\orcid}}\hspace{1mm}Daniele Cirulli}
\author[3]{\href{https://orcid.org/0009-0000-8301-136X}{\usebox{\orcid}}\hspace{1mm}Edoardo Di Martino}
\author[4]{\href{https://orcid.org/0009-0003-7169-0774}{\usebox{\orcid}}\hspace{1mm}Jacopo Nudo}
\author[5]{\href{https://orcid.org/0000-0002-6227-6024}{\usebox{\orcid}}\hspace{1mm}Maria Sahakyan}
\author[3]{\href{https://orcid.org/0009-0003-1024-3735}{\usebox{\orcid}}\hspace{1mm}Emanuele Sangiorgio}
\author[6]{\href{https://orcid.org/0009-0006-6039-1594}{\usebox{\orcid}}\hspace{1mm}Arnaldo Santoro}
\author[3]{\href{https://orcid.org/0009-0004-0682-8578}{\usebox{\orcid}}\hspace{1mm}Simon Zollo}
\author[1]{\href{https://orcid.org/0000-0001-6859-0391}{\usebox{\orcid}}\hspace{1mm}Alessandro Galeazzi}
\author[4]{\href{https://orcid.org/0000-0003-4335-7328}{\usebox{\orcid}}\hspace{1mm}Niccolò Di Marco}

\affil[1]{Department of Mathematics, University of Padova, Padova, 35121 \\ \texttt{eleonora.amadori@studenti.unipd.it},
\texttt{alessando.galeazzi@unipd.it}}
\affil[2]{Department of Physics and INFN, University of Rome ``Tor Vergata'', Rome, 00133, Italy \\ ``Enrico Fermi'' Research Center, Rome, 00184, Italy \\ \texttt{daniele.cirulli@cref.it}}
\affil[3]{Department of Social Sciences and Economics, Sapienza University of Rome, Rome, 00185 \\ \texttt{edoardo.dimartino@uniroma1.it}, \texttt{emanuele.sangiorgio@uniroma1.it}, \texttt{simon.zollo@uniroma1.it}}
\affil[4]{Department of Computer Science, Sapienza University of Rome, Rome, 00161 \\ \texttt{jacopo.nudo@uniroma1.it}, \texttt{niccolo.dimarco@uniroma1.it}}
\affil[5]{Social Science Division, New York University Abu Dhabi, Abu Dhabi, 129188 \\ \texttt{ms13502@nyu.edu}}
\affil[6]{Department of Environmental Sciences, Informatics, and Statistics, Ca' Foscari University of Venice, Venezia, 30123 \\ \texttt{arnaldo.santoro@unive.it}} 

\hypersetup{
pdftitle={A template for the arxiv style},
pdfsubject={q-bio.NC, q-bio.QM},
pdfauthor={David S.~Hippocampus, Elias D.~Striatum},
pdfkeywords={First keyword, Second keyword, More},
}

\begin{document}
\maketitle


\begin{abstract}
Language is a fundamental aspect of human societies, continuously evolving in response to various stimuli, including societal changes and intercultural interactions. Technological advancements have profoundly transformed communication, with social media emerging as a pivotal force that merges entertainment-driven content with complex social dynamics. As these platforms reshape public discourse, analyzing the linguistic features of user-generated content is essential to understanding their broader societal impact.
In this paper, we examine the linguistic complexity of content produced by influential users on Twitter across three globally significant and contested topics: COVID-19, COP26, and the Russia–Ukraine war. By combining multiple measures of textual complexity, we assess how language use varies along four key dimensions: account type, political leaning, content reliability, and sentiment.
Our analysis reveals significant differences across all four axes, including variations in language complexity between individuals and organizations, between profiles with sided versus moderate political views, and between those associated with higher versus lower reliability scores. Additionally, profiles producing more negative and offensive content tend to use more complex language, with users sharing similar political stances and reliability levels converging toward a common jargon.
Our findings offer new insights into the sociolinguistic dynamics of digital platforms and contribute to a deeper understanding of how language reflects ideological and social structures in online spaces.
\end{abstract}


\section{Introduction}
\label{sec:introduction}
Language has played a pivotal role in the development of human society, functioning as a fundamental medium for the transmission of knowledge and the coordination of social interaction~\cite{crystal2018cambridge}. Over time, languages have evolved in response to several factors, such as changing socio-political conditions and intercultural contact, which may result in lexical borrowing, code-switching, and, in some cases, the formation of new dialects or creole languages~\cite{thomason2023language}. As technological and socio-political landscapes continue to change, the nature and intensity of intercultural interactions have also changed, particularly with the advent of digital communication technologies, which have dramatically increased the volume, speed, and global reach of such exchanges~\cite{androutsopoulos2014mediatization, sangiorgio2025evaluating}.
In this landscape, the rise of social media platforms plays a pivotal role, profoundly impacting the way people connect, communicate, and access information.
With their increasing popularity, these platforms have emerged as a key data source to analyze public discourse, information diffusion, and social dynamics on a large scale. Their unique structure, comprising short user-generated messages, hashtags, mentions, and share mechanisms, has allowed researchers to study a wide range of troubling phenomena such as increased polarization~\cite{flaxman2016filter, cinelli2021echo, nyhan2023like, tucker2018social, falkenberg2022growing,benanti2020se}, the spread of misinformation~\cite{delvicario2016misinformation, baillon2023asymmetrical}, and the amplification of hate speech~\cite{CASTANOPULGARIN2021101608, siegel2020online}. Unlike traditional media, Social Media Platforms offer immediacy, broad reach, and high variability in sentiment, tone, and engagement, making it a rich source for large-scale behavioral analysis~\cite{gilani2019large, avalle2024persistent, di2024ideological}, natural language processing~\cite{hasan2019sentiment, muller2023covid}, and public opinion research ~\cite{zollo2025inference}. 
Furthermore, previous studies have utilized social media data to investigate not only the content of online conversations but also the behavioral patterns and influence dynamics of users~\cite{lewandowsky2020using, enli2017twitter}. 

Within this context, increasing interest is growing in analyzing the linguistic features of user-generated content~\cite{greaves2013sentiment, xu2019sentiment, alrumaih2020sentiment, volkova2015inferring}.
These studies offer important insights into the complex social dynamics of communication in digital platforms~\cite{perc2012evolution}. Nevertheless, accurately measuring language complexity remains a challenging task, particularly in specialized domains that go beyond traditional linguistic and communicative-discursive frameworks—most notably in psycho-cognitive areas such as reading comprehension and information processing~\cite{smith1941measurement, tschirner2004breadth}.
Although extensive research in the educational field to examine the impact of vocabulary size and highlight individual differences in lexical knowledge has been conducted~\cite{milton2013vocabulary}, the evolution of these dynamics in the digital era remains largely unexplored. 

Driven by concerns that internet may corrupt traditional writing conventions and face-to-face communication \cite{baron2010always, mcculloch2020because, zappavigna2012discourse, baron2015words}, interest in linguistic research on social media has been recently increasing. 
Notably, recent studies exhibit patterns of linguistic simplification~\cite{dimarco2024vocabulary,desiderio2025nexus} in the comments published in social media, while other researches measure the same process of simplification in different domains~\cite{parada2024song,di2025decoding}. 
Despite recent efforts, a fundamental question remains unanswered: do users with different stances or characteristics employ distinct vocabulary and linguistic properties?

In this study, we leverage a large-scale dataset of tweets posted by influencers on Twitter/X, covering three major topics: COVID-19, COP26 and the Ukraine war. In addition, we collect detailed information on each influencer's stance and social characteristics, allowing for a more in-depth analysis.

Using this dataset, we apply multiple measures of textual complexity to the tweets, extracting a range of linguistic features. Moreover, we apply tools from Network Science to understand how similar influencers are in terms of used language.

Our results reveal differences in the complexity  of the language used by individual or organizations, as well as an effect of political stances and levels of reliability. 
Moreover, noxious semantic traits of users' language such as offensiveness and negative sentiment result in an higher complexity of the users' vocabulary. 

Finally, the semantic network of influencers reveals distinct clusters reflecting a convergence between language similarity, political bias, and reliability.

By unveiling the linguistic and ideological structures that underlie language complexity among key actors in the digital space, our study sheds light on the mechanisms driving opinion formation and content dissemination in contemporary online ecosystems—ultimately contributing to a deeper understanding of how public discourse is shaped in the digital age.

\section{Methods}
\label{sec:methods}

\subsection{Data Collection}
\label{subsec:data_collection}
All data used in this work were collected via the former Twitter API for Academic Research, using search queries specific to each dataset. These datasets were originally compiled and curated as part of a previous study~\cite{loru2024sets}, from which we built upon for our analysis. We briefly summarize how the datasets are gathered.

Tweets and user information related to the COP26 debate were collected using the keyword ``cop26'', spanning the period from 1 June 2021 to 14 November 2021

Data about COVID-19 were collected between 1 January 2020 and 30 April 2021 using keyword searches that included: ``vaccin", ``dose", ``pharma", ``immun", ``no-vax", ``novax", ``pro-vax", ``provax", ``antivax", and ``anti-vax". 

Tweets about Ukraine were collected by tracking specific keywords related to the Russia-Ukraine conflict, spanning the period from 22 February 2022 to 17 February 2023. The tweet IDs were obtained from a publicly available dataset~\cite{chen2023tweets}, from which a random sample comprising 25\% of the available IDs was downloaded.

For each dataset, a set of ``influencers'' who played a prominent role in shaping the online debate was identified. Starting from the top 1\% most retweeted accounts, those also ranked in the top $50\%$ of original tweet producers (excluding replies, quotes, and retweets) were selected to ensure consistent and active participation. 

Influencers were manually labelled using questionnaires by the authors of Loru et al~\cite{loru2024sets}. Each account was assigned to one of six categories based on their social, political, or professional role: Activist, International Organization/NGO, Media, Politics, Private Individual, or Other. Influencers were annotated independently for each topic, and only if they appeared among the selected accounts for that particular dataset.

In the current study, we focus on english tweets published by the influencers. A data breakdown of the resulting dataset is shown in Table~\ref{tab:data_breakdown}.

We clarify that these datasets have been selected as they pertain to topics that have been the subject of significant debate in recent years.

\begin{table}[!ht]
\caption{Breakdown of the datasets used in the study. For each dataset, we report the number of tweets, the number of unique users, and the time span covered in the analysis.}
\label{tab:data_breakdown}
\centering
\begin{tabular}{lllc}
\hline
\textbf{Event} & \textbf{N. of Tweets} & \textbf{N. of unique users} & \textbf{Time-span}      \\ \hline
COVID-19       & $607192$                & $1488$                        & $01/01/2020 - 30/04/2021$ \\ 
COP26          & $135636$                & $561$                         & $01/06/2021 - 14/11/2021$ \\ 
Ukraine        & $916955$                & $967$                         & $22/02/2022 - 17/02/2023$ \\ \hline
\textbf{Total} & \textbf{1659783}      & \textbf{3016}               & \textbf{}               \\ 
\end{tabular}
\end{table}

\subsection{Text Pre-processing}

All tweets were preprocessed to keep only the significant part of the texts and avoid spurious results. All the tweets were also cleaned of emojis, hashtag, tag, and URL.
To compute $K$-complexity and vocabulary size, we concatenate all texts published by an influencer and use the {\it quanteda} package in R to apply the following pre-processing steps:

\begin{itemize}
    \item We tokenize the texts and remove punctuation and English stopwords;
    \item We reduce the resulting tokens to their root form by applying a stemmer;
    \item We set each token to lowercase.
\end{itemize}

In this way, we obtain a set of tokens (words) and types (unique words) for each influencer.

Note that, since a text may remain empty after these steps (for example, in the case it contains only tags or hashtags), we removed all users containing 0 tokens.

\subsection{Labeling Users’ Political Stance and Factual Reliability}
To classify users according to political stance and the reliability of the content they shared, we employ Google’s Gemini large language model, in line with recent studies demonstrating the effectiveness of large language models for social media annotation tasks~\cite{renault2025republicans, gilardi2023chatgpt}. Users are categorized as Left, Center, or Right leaning in terms of political orientation, and as either Reliable or Questionable with respect to the factual credibility of the content they disseminated.
To ensure robustness, we validate the model-generated labels against external assessments from MediaBias/FactCheck (MBFC, see \url{https://mediabiasfactcheck.com}), a widely used independent news rating agency that provides political bias and reliability labels for news outlets and other information producers~\cite{ cinelli2021echo, stefanov2020predicting, flamino2023political}. Although the overlap between our dataset and MBFC is limited (56 matched users), we compute Cohen’s Kappa coefficient to assess the agreement between the two labeling systems. Results indicate a level of agreement from moderate to substantial for both political stance ($\kappa=0.58$) and content reliability ($\kappa=0.75$), suggesting that our automated labeling procedure is both reliable and consistent with established external benchmarks. The corresponding confusion matrices are reported in Figure~\ref{fig:fig_confusion_matrix_1_SI} and Figure~\ref{fig:fig_confusion_matrix_2_SI}  of the Supplementary Information appendix.

\subsection{Sentiment and Offensive Language Classification}

Transformer-based language models consistently demonstrate excellence performance in tweet classification tasks due to their ability to capture contextual, informal, and affective nuances present in social media text~\cite{barbieri2020tweeteval}. One of the most widely adopted models is RoBERTa (Robustly Optimized BERT Pre-Training Approach), a variant of BERT that removes the next-sentence prediction objective, trains with significantly more data, uses larger mini-batches and learning rates, and dynamically changes the masking pattern during training to improve the model’s ability to learn contextual representations~\cite{liu2019roberta}. Its architecture makes it particularly well-suited for handling the lexical variability, slang, and conversational structure of social media content. 

For sentiment analysis, RoBERTa and its variations achieve strong performance across diverse domains~\cite{gaurav2024xlm, barreto2023sentiment, krishnamoorthy2024analyzing, rawther2023transformer, 
tan2022roberta}. Comparative evaluations show that RoBERTa and BERTweet outperform earlier transformer architectures when adapted to in-domain Twitter content~\cite{nguyen2020bertweet}. In the area of offensive language detection, RoBERTa-based models demonstrate state-of-the-art performance across multiple benchmark evaluations. Barbieri et al.~\cite{barbieri-etal-2020-tweeteval} introduced a RoBERTa model fine-tuned on a standardized suite of Twitter-based classification tasks, showing superior results in offensive language detection relative to prior models. Studies confirm the effectiveness of RoBERTa variants in identifying offensive, abusive, or toxic language in multilingual and event-driven Twitter datasets, consistently reporting high accuracy and improved generalization across domains~\cite{kaati2022machine, wiedemann2020uhh}.

Given the effectiveness and scalability of RoBERTa-based models in both sentiment and offensive language classification tasks, we selected domain-specific variants fine-tuned on Twitter data for our analysis. To fully exploit their capabilities and ensure reliable inference, we applied a structured pre-processing pipeline to align the raw tweet content with the models’ input requirements.

In particular, to ensure the compatibility of Twitter content with RoBERTa-based models and to minimize noise, we first convert all tweets into a standardized textual format, addressing issues of encoding, syntactic artifacts, and non-standard symbols commonly found in user-generated content. Hypertext markup, external links, and long isolated numeric sequences are removed, as they offer limited linguistic value for sentiment or toxicity classification. User handles (i.e. ``@" mentions) are replaced with a standardized placeholder to reduce lexical variability while preserving conversational structure. Emojis are converted into textual descriptors to preserve their affective content in a form compatible with the model’s tokenizer. Punctuation and typographic inconsistencies, including non-standard quotation marks and excessive whitespace, are normalized, and non-ASCII characters are excluded to prevent tokenization errors. Tweets are also screened for minimal linguistic content: entries consisting exclusively of user mentions, retweet shells without commentary, or only hashtags are excluded. Furthermore, tweets dominated by non-verbal tokens (e.g., emojis or symbols), are flagged for quality control. These pre-processing steps ensure that the input data conforms to the structural and lexical conventions of the pre-trained RoBERTa models, thereby enhancing the robustness of downstream inference.

Following pre-processing, we apply two transformer-based language models to classify each tweet along distinct dimensions: sentiment and offensive language. Both models are based on the RoBERTa architecture and were pre-trained on large-scale Twitter corpora. The sentiment classification model categorizes tweets into three classes: positive, neutral and negative~\cite{loureiro-etal-2022-timelms} while the offensive language classifier distinguishes between offensive and non-offensive tweets~\cite{barbieri2020tweeteval}. Computation is performed on a GPU-enabled environment to accelerate inference, and the resulting predictions are used in subsequent analyses of linguistic patterns and discourse dynamics across events.

In total, we process 1,359,793 unique tweets authored by 3,016 distinct users. To enable user-level analysis of sentiment and offensive behavior, we aggregate the model predictions across all tweets associated with each user. Specifically, we calculate the proportion of tweets labeled as negative to represent a user's overall negativity score, and similarly compute the percentage of tweets classified as offensive to quantify their offensiveness level. We rely on these aggregated metrics to analyze patterns in how individual users express themselves on Twitter and perform a systematic comparisons between different groups. 




\subsection{Textual Complexity}
Text complexity can be analyzed from multiple perspectives, and the academic literature offers a wide range of metrics for this purpose~\cite{jensen2009indicators, zipf2013psycho, tanaka2015computational, renyi1961measures, dugast1979vocabulaire, tweedie1998variable}. Following a thorough review of the existing measures and related research, we select two specific metrics that capture different, complementary dimensions of textual complexity: Yule’s $K$ measure of lexical richness, gzip-based compression complexity and Flesch readability score.\\
We remove, on the raw text of tweets, each ``@'', ``\#'', and emojis found in them, to avoid any distortion of the results.

These three measures provide a relatively orthogonal view, allowing for a more comprehensive assessment of three different text characteristic: lexical complexity, repeatitivity and easiness of reading.

In more detail, 
Yule’s $K$-complexity~\cite{yule2014statistical} quantifies how repetitive or diverse the vocabulary is, and it is considered to be largely independent of the overall length of a text. In detail, for a text of length $N$ containing a $V$ unique words, Yule’s $K$ is defined as $$K = 10^4 \cdot \left[-\frac{1}{N} + \sum_{i = 1}^{V} V(i,N) \left(\frac{i}{N}\right) ^2 \right],$$ where $V(i,N)$ represents the number of words that occur $i$ times. The measure has a theoretical lower bound of 0, which is achieved only when all words in the text are distinct (e.g., $V(1,N)=N$). In general, as $K$ increases, the lexical richness of the text decreases, although there is no theoretical upper bound for the measure.

To assess the repetitiveness of a text, we adopt an approach similar to previous studies~\cite{parada2024song,desiderio2023recurring, dimarco2024vocabulary}. 

This approach estimates the compressibility of concatenated tweets for each user, with the underlying assumption that more repetitive or predictable texts compress better than complex or diverse ones.

The procedure involves measuring the size in bytes of the original text (encoded in UTF-8) and then compressing the same text using the G-Zip algorithm. The compressed size is also recorded. From these measurements, we compute the \textit{compression ratio} defined as the size of the compressed text divided by the size of the original text:

\begin{equation}
    gzip = \frac{s_{compressed}}{s_{raw}},
\end{equation}

where $s_{raw}$ represents the size of the raw text and $s_{compressed}$ the size of the compressed text. 

Lower compression ratios indicate higher redundancy and lower complexity, whereas higher ratios suggest richer and less predictable language use. This metric complements traditional readability scores by capturing structural and informational properties of the text.

To evaluate the readibility of tweets, we also consider the \textit{Flesch Index}. This metric is commonly used to assess the readability of English texts, providing a numerical score based on sentence length and syllable density. The formula is defined as follows:

\begin{equation}
\text{Flesch Index} = 206.835 - 1.015 \cdot \left( \frac{\text{\# words}}{\text{\# of sentences}} \right) - 84.6 \cdot \left( \frac{\text{\# of syllables}}{\text{\# of words}} \right)
\end{equation}

Higher scores indicate texts that are easier to read (with standard journalistic English typically scoring between 60 and 70), whereas lower or negative scores reflect more complex or less natural language structures. 

\subsection{Statistically Validated Projections for Bipartite Networks}
\label{subsec:statsval}

In this section, we discuss our procedure to construct the influencers network.
First, we build a weighted bipartite network \(W=\{w_{i\alpha}\}\) between influencers \(i\) and content‐types \(\alpha\), where \(w_{i\alpha}\) measures how extensively influencer \(i\) uses type \(\alpha\). 
To focus on the most relevant links, we binarize \(W\) via the Revealed Comparative Advantage (RCA) filter~\cite{rcabalassa1965trade}:
\begin{equation}
\label{eq:RCA}
\mathrm{RCA}_{i\alpha}
= 
\frac{\displaystyle\frac{w_{i\alpha}}{\sum_\beta w_{i\beta}}}
     {\displaystyle\frac{\sum_j w_{j\alpha}}{\sum_{j,\beta} w_{j\beta}}}
\quad\Longrightarrow\quad
m_{i\alpha} =
\begin{cases}
1, & \text{if }\mathrm{RCA}_{i\alpha} > 1,\\
0, & \text{otherwise,}
\end{cases}
\end{equation}
\noindent
where the binary matrix \(M=\{m_{i\alpha}\}\) retains only links exceeding the expected global usage.\\
\noindent
Next, to project \(M\) onto the influencer layer while correcting for degree bias—i.e., the tendency of high‐degree influencers to exhibit spurious overlaps—and filtering out random noise, we employ the Bipartite Configuration Model (BiCM) \cite{Sar_2017_monop,Cimini2019}, a maximum-entropy null model for bipartite networks.
\noindent
This approach constructs an ensemble of networks \(\mathcal{G}\) that is maximally random except that it preserves the degree sequences: \(k_i\) for each influencer \(i\) and \(\kappa_\alpha\) for each type \(\alpha\).
\noindent
We then derive, for each influencer pair \((i,j)\), the null distribution of their shared neighbors and compute the corresponding \(p\)\nobreakdash-value.\\
\noindent
To this end, we perform a constrained maximization of the Shannon entropy, which quantifies the uncertainty in the network’s configuration:

\begin{equation}
S = -\sum_{G\in\mathcal{G}} P(G)\,\ln P(G),
\end{equation}
\noindent
The equation \(\langle C\rangle = \sum_{G}P(G)\,C(G)\) determines the ensemble average of the constraint. 
\noindent
Introducing Lagrange multipliers \(\{\theta_i\}\) and \(\{t_\alpha\}\) associated with these constraints, we define the Hamiltonian
$H(G;\{\theta_i,t_\alpha\}) = \sum_i \theta_i\,k_i(G) + \sum_\alpha t_\alpha\,\kappa_\alpha(G)$.\\
\noindent
Maximizing the entropy subject to constraints on the degrees of influencers \(k_i\) and types \(\kappa_\alpha\) yields the graph probability distribution

\begin{equation}
P(G \mid \{\theta_i,t_\alpha\})
= \frac{e^{-H(G;\{\theta_i,t_\alpha\})}}{Z(\{\theta_i,t_\alpha\})}\,,
\end{equation}

\noindent
where $Z(\vec{\theta}) = \sum_{G \in \mathcal{G}} e^{-H(G, \vec{\theta})}$ is the partition function over the ensemble.\\
\noindent
Solving the model, requiring that the ensemble averages of the constraints match their empirical values in \(G^*\), yields the independent link probabilities \(p_{i\alpha}\) \cite{Sar_2017_monop}.\\
\noindent
We then compute, for each influencer pair \((i,j)\), the null distribution of their shared neighbors under independent Bernoulli trials and obtain the corresponding \(p\)\nobreakdash-value for each projected link.\\
\noindent
Finally, we apply a false discovery rate (FDR) multiple‐testing correction \cite{fdrBenjamini1995}, retaining only those influencer pairs with statistically significant co‐occurrences in the validated projection.\\
\noindent
All statistical validations were implemented using the software package described in Vallarano et al. \cite{Vallarano2021}.\\
\noindent
Network metrics were computed as detailed in the Supporting Information appendix. Community detection was performed by maximizing modularity with the Louvain algorithm \cite{NX}, and network layouts for visualization were generated using Gephi \cite{gephi2009}.

\section{Results}
\label{sec:results}

\subsection{Summary Statistics of Influencer Activity and Vocabulary Size}
\label{subsec:summary_stat}
\begin{figure}[!t]
       \centering
    \includegraphics[width=\textwidth]{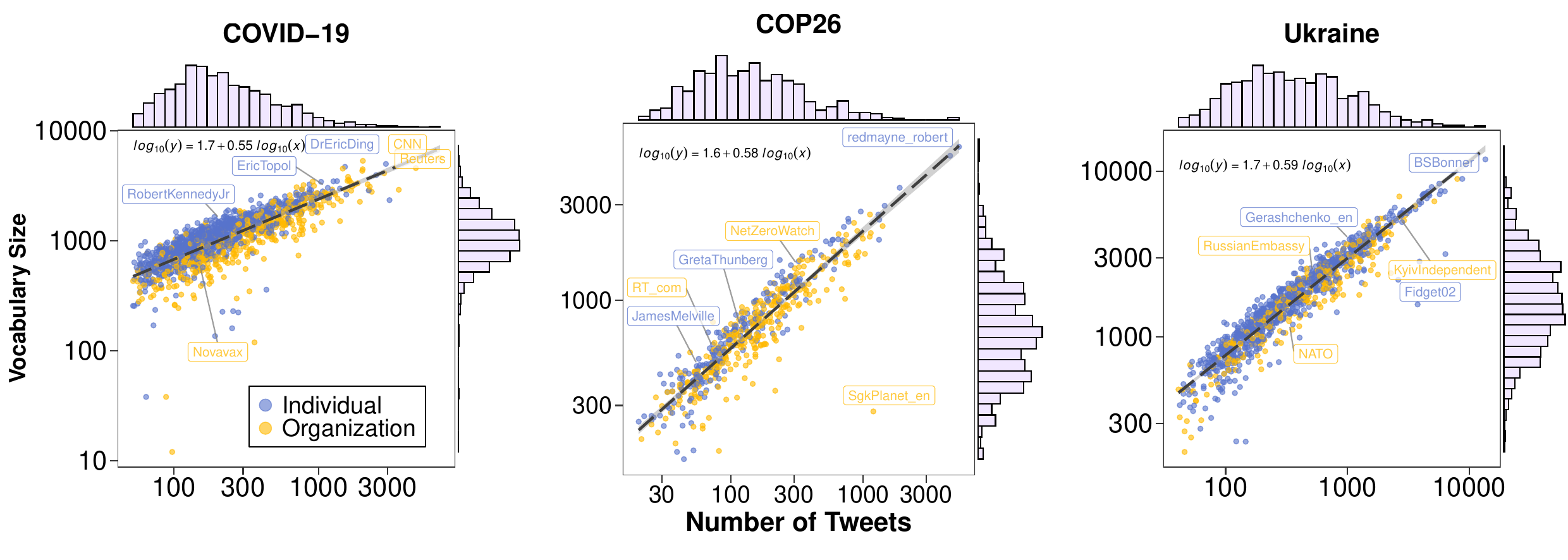}
        \caption{Bivariate distribution of the number of tweets and the vocabulary size for each influencer considered in the analysis. The color indicates whether the influencer is an individual or an organization. Notably, the dataset reveals distinct clusters for each category. Regression lines obtained through Ordinary Least Squares (OLS) are also shown. Log-transformed marginal distributions are also shown on the axes.}
        \label{fig:fig_1}
\end{figure}

We begin our analysis by presenting an overview of the three datasets. Figure ~\ref{fig:fig_1} illustrates the relationship between the number of tweets and the size of the vocabulary (i.e., the number of word types) for each influencer. The marginal distributions of tweet count and vocabulary size are also shown, both exhibiting a heavy-tailed pattern that indicates that most influencers produced relatively few tweets and employed a rather limited vocabulary.\\

Given that the two variables are log-transformed, the regression line reveals a power-function relationship between the number of posts published and the vocabulary size.

Interestingly, we observe consistent trends across topics, characterized by comparable slope values, a notable regularity between datasets.

Across all topics, the most active influencers tend to use a wider variety of types (i.e., unique words used). 
Notable examples include $@Reuters$ (organization) for COVID-19, $@redmayne\_robert$ (individual) for COP26, and $@BSBonner$ (individual) for the Russia-Ukrainian event. Interestingly, some influencers behave differently from others. For instance, $@SgkPlanet\_en$ (activist organization) during the COP26 event and $@Fidget02$ (private individual) during the Russia-Ukraine war published a large number of tweets but used a significantly limited vocabulary. These patterns highlight the importance of considering both activity level and linguistic diversity when analyzing influence and communication strategies across different events.
Furthermore, the plots reveal the presence of distinct clusters for individual and organizational influencers, suggesting different vocabulary characteristics. \\
Overall, this first analysis provides a descriptive understanding of influencers' activity and language use, revealing consistent patterns between vocabulary size and tweet volume across topics, as well as systematic differences between individual and organizational accounts.

\subsection{Complexity Analysis}
\label{subsec:complexity}
We then focus on the analysis of text complexity, comparing complexity measures across five different spectra: account type, political leaning, content reliability, users' negativity of sentiment, and users' offensiveness. 

Lexical complexity is quantified using Yule's $K$, while repeatitivity and readability aspects are measured using the {\it gzip} compression ratio and the Flesch Reading Ease Index, respectively, as detailed in the Methods section. We report only the results for Yule's $K$ in the main text, while  results for the latter two are illustrated in Figure ~\ref{fig:fig_gzip_SI} and Figure ~\ref{fig:fig_flesch_SI} in the Supplementary Information appendix.

\begin{figure}[!t]
       \centering
       \includegraphics[width=.75\textwidth]{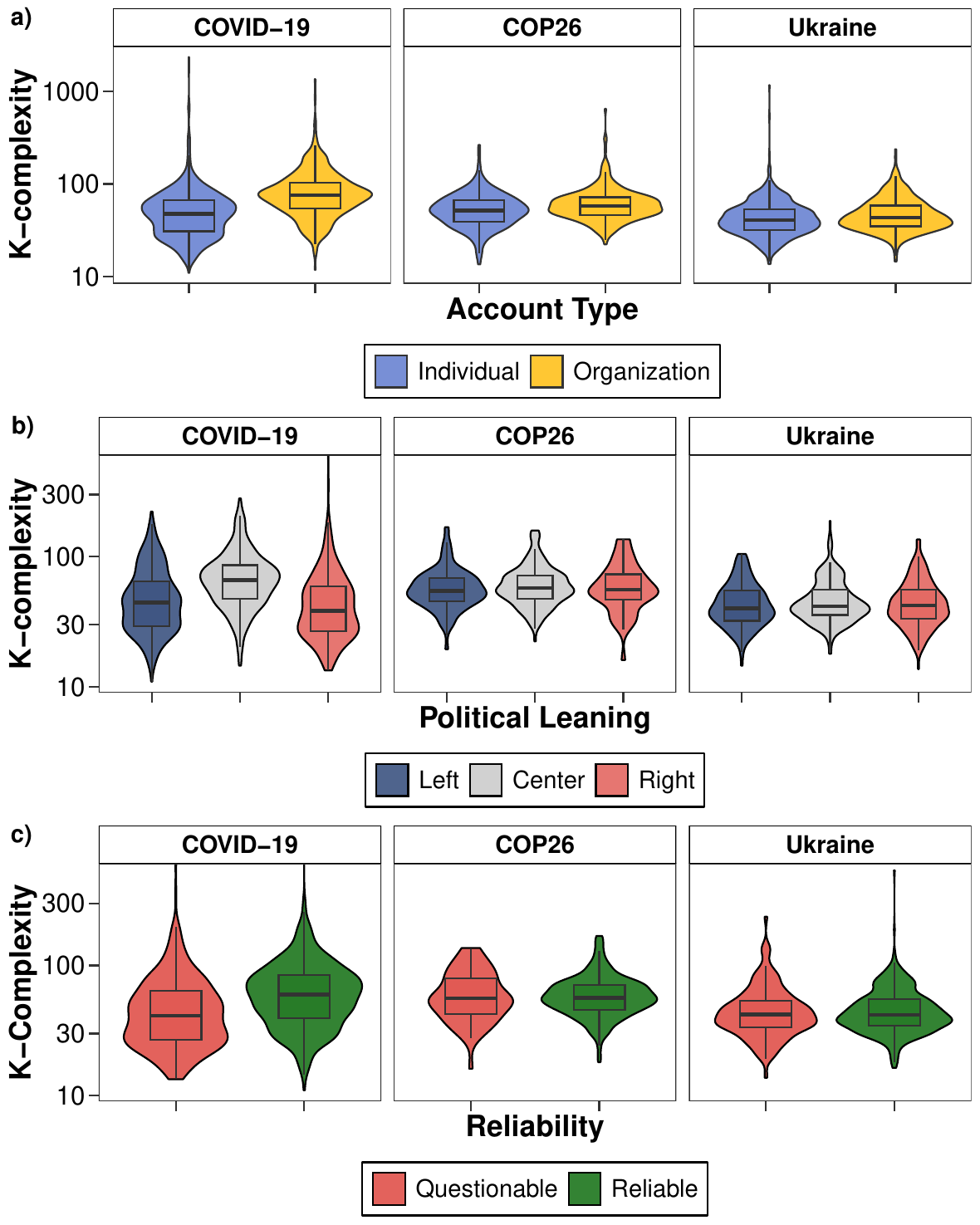}
        \caption{$K$-complexity distributions according to a) the account type of the influencers (Individual or Organization), b) the political leaning of the influencer (Left, Center, Right), c) the reliability of the influencer (Questionable or Reliable).
        }
        \label{fig:fig_2}
\end{figure}

\paragraph{Category, political stance and reliability}
Figure ~\ref{fig:fig_2} shows the $K-$complexity distributions according to the account type (individual vs.\ organization), political stance, and reliability of the account, determined as explained in the Methods section.

Notably, as shown in Panel~a), individuals exhibit statistically significantly lower $K$-complexity scores, showing how their language appears to be more complex than that used by organizations. These results are persistent and significant across all three datasets as confirmed by the Kruskal-Wallis test, the results of which are reported in Table ~\ref{tab:tab_test} of the Supporting Information appendix. 

In Panel~b), it can be observed how profiles expressing a lower political bias (Center) display higher values of $K$-complexity scores, while both left and right wing profiles tend to exhibit lower values of the metric, and so showing a higher complexity of vocabulary used. 
This is possibly indicating that those who are more engaged in writing messages along partisan lines tend to use a vocabulary greater in size and complexity.

However, the Kruskal-Wallis tests highlight that only the COVID-19 dataset displays significant differences based on the political leaning of influencers, while the COP26 and the Ukraine datasets show no statistically significant differences.

Finally, Panel~c) shows that accounts flagged as questionable in terms of content reliability tend to use more lexically complex language in the COVID-19 dataset. For the COP26 and Ukraine datasets, the differences are not significant. 

However there is a degree of variation in results across the three datasets, we show in Figure ~\ref{fig:fig_gzip_SI} and ~\ref{fig:fig_flesch_SI} of the Supplementary Information appendix that, when employing {\it gzip} or \textit{Flesch Index} as a measure of redundancy of language and readability, the patterns discussed above are shown to be persistent across the datasets.

\paragraph{Offensiveness and Sentiment}

To investigate the connection between linguistic complexity and user behavior on Twitter during significant global events, we examine the association between Yule's $K$-complexity and aggregated measures of offensiveness and negative sentiment at the user level. The users were grouped into four categories, \textit{low, medium, high and very high}, according to the quartile of the distribution of their respective scores. 

\begin{figure}[!t]
       \centering
    \includegraphics[width=0.9\textwidth]{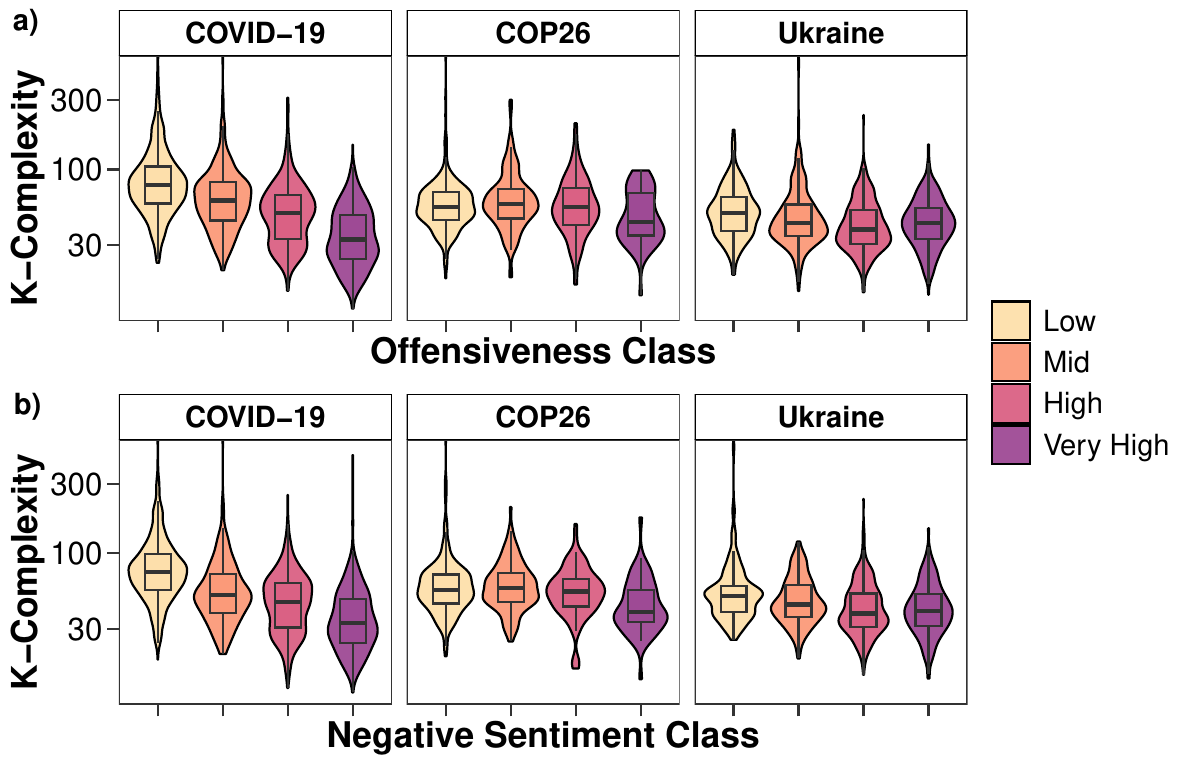}
        \caption{$K$-complexity distributions by (a) user offensiveness class and (b) negative sentiment class.
    The offensiveness class is defined based on how frequently a user posts offensive tweets, while the negative sentiment class is determined by the average negativity score of the user's tweets.}
        \label{fig:fig_3}
\end{figure}

Figure~\ref{fig:fig_3} shows the relationship between $K$-complexity, offensiveness, and negativity at the user-level. 

As we can observe from both Panel~a) and b), users in higher classes tend to exhibit lower $K$-complexity scores, indicating the use of more complex language. 
This inverse relationship is most pronounced in the context of COVID-19, but the same trend is observed for COP26 and the Russia-Ukraine conflict, as confirmed by Figure~\ref{fig:fig_gzip_SI} and Figure~\ref{fig:fig_flesch_SI} in the Supporting Information appendix. 
Overall, the pattern remains consistent across all three events. This suggests that users who are more engaged in writing content that carries negative or offensive weight tend to express their perspectives with greater linguistic complexity.

\subsection{Analysis of Influencer Networks}

\begin{figure}[!t]
       \centering
    \includegraphics[width=0.9\textwidth]{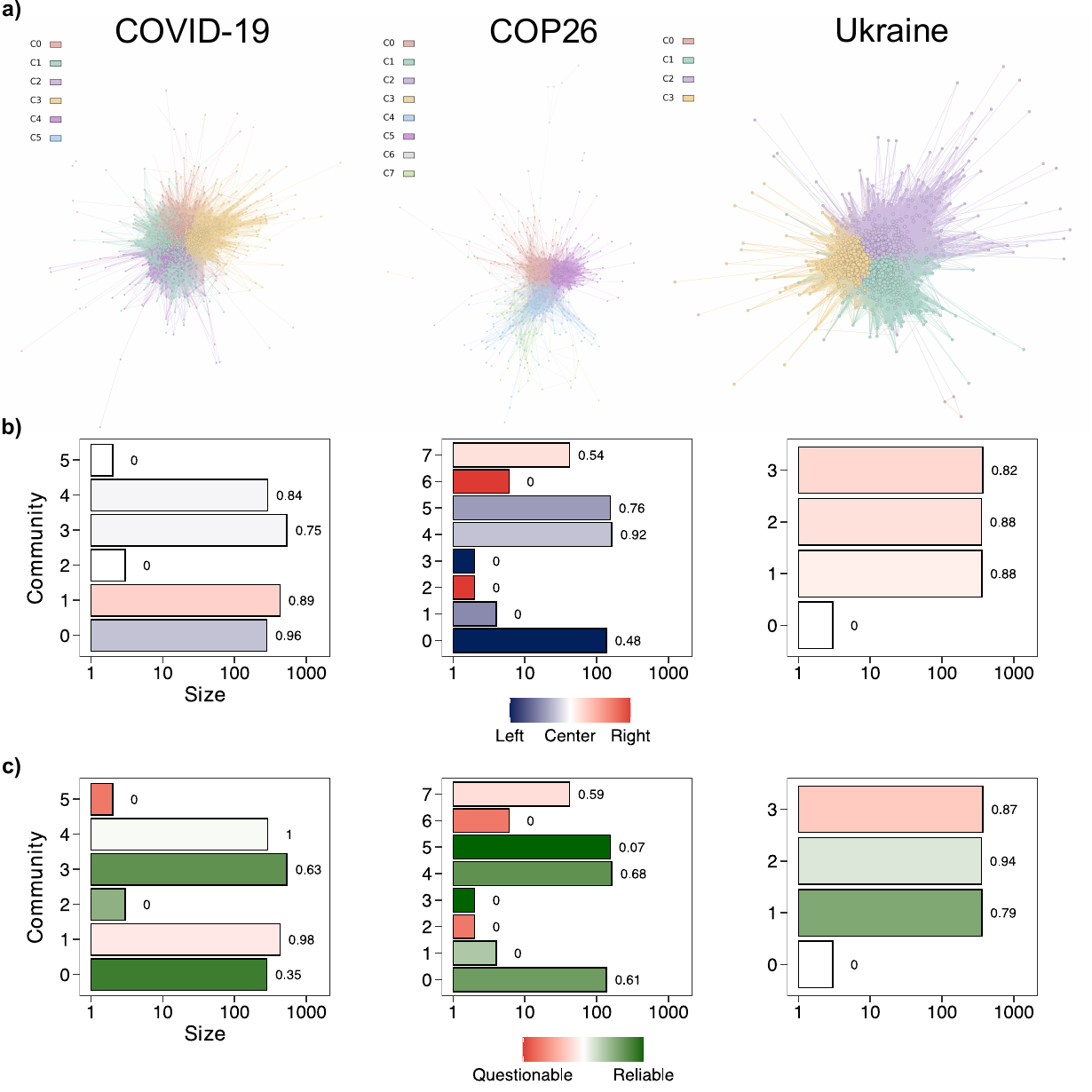}
        \caption{$(a)$ Visualization of the influencers' networks for the three datasets. In panels $(b)$ and $(c)$, Size represent the number of influencers in each community, colors represent the political leaning ($b$) or the reliability label ($c$) of the communities detected using the Louvain algorithm. For each community, the number next to the bar represents the value of the Shannon Entropy calculated on the relative distribution of the political leaning or reliability label.}
        \label{fig:fig_4}
\end{figure}

Building on our earlier findings, we aim to investigate whether distinct groups of influencers use different lexical choices. To this end, for each of the three datasets we construct a weighted bipartite network linking ``Influencers'' to the ``Types'', defined as the unique words they use.
First, we binarize the bipartite network using the RCA filter (see Methods \ref{subsec:statsval}). 
Next, we perform a statistical validation based on the Bipartite Configuration Model (BiCM) (see Methods \ref{subsec:statsval}) when projecting the bipartite network onto the influencer layer. 
This step filters out random noise, yielding a one‐mode network in which edges represent only statistically significant co‐occurrences of influencers sharing types.
Tables ~\ref{tab:bipartite_metrics} and~\ref{tab:network_metrics} summarize the original bipartite networks and their statistically validated projections (see Supplementary Information). 
Although the bipartite graphs are large, the BiCM-based validation filters out random noise (see Methods \ref{subsec:statsval}), eliminating several influencers in the projection (i.e. those who share no types at a statistically significant level; see ~\ref{tab:network_metrics} in Supplementary Information).
We then detect communities in each co‐occurrence network using the Louvain algorithm by maximizing modularity (see Table~\ref{tab:network_metrics}).

In Figure~\ref{fig:fig_4}a), we represent the three networks, highlighting the communities we found. 

To understand if these communities are defined by a clear political orientation or reliability labels, we compute the average political leaning and reliability scores of the users they comprise, along with the Shannon Entropy of these distributions to assess the cluster's heterogeneity~\cite{Virtanen2020SciPy}. 
In doing so, we remove users with unknown political bias or unknown reliability labels to normalize the data. 

\noindent

The results, shown in Panel~b) and c) of Figure~\ref{fig:fig_4}, suggest that communities can be characterized by the political stance or reliability of their influencers. 
The COP26 dataset seems to be better clustered through the political dimension than Covid and Ukraine.
On the other hand, all three datasets exhibit nice clustering according to the reliability score.

Notably, this indicates that influencers sharing similar point of view both from political and reliability are more likely to use similar words.

Finally, we note how the alignment between the cluster community structure and the political / reliability dimensions is strongest in the dataset about COVID-19 and COP26 networks, precisely those with the highest modularity scores (reported in Tab. \ref{tab:network_metrics}), probably reflecting the more science-driven nature of these topics (e.g., vaccines, climate science) compared to debates rooted in politics. This could suggest that scientific topics tend to distinguish more the language used by actors with opposite opinions.


\section{Discussion}
\label{sec:discussion}
We investigate the complexity of language on social media by analyzing a set of influential Twitter/X users across three highly debated topics: COVID-19, COP26, and the Russia-Ukraine war. We study several aspects that can influence language complexity, measuring its variations along four dimensions: account type, political stance, reliability, and sentiment. 
To this end, we employ a diverse set of metrics and methodologies, including $K$-complexity, $gzip$ compression, Flesch readability index, sentiment analysis, and network analysis.

Our results reveal several noteworthy insights into how influencers communicate on social media.
First, we identify a consistent polynomial relationship between the size of an influencer’s vocabulary and the volume of content published across all three topics.
We also find that individuals tend to use more complex language than organizations, possibly due to greater freedom in expressing personal opinions.

In addition, according to two out of three employed complexity measures, language complexity varies with political stance and reliability: users with stronger partisan alignments exhibit higher complexity than politically neutral users, and accounts labeled as questionable show greater complexity than those deemed reliable.
Further, NLP analyses indicate that users who frequently post negative or offensive content also tend to use more complex language.
Finally, our network-based analysis highlights how users with similar political stances or reliability labels are more likely to produce linguistically similar content.
Our results suggest that users with greater involvement, even of a different nature, tend to use more complex language in online debates. This dynamic is consistent across the three global debates.
Moreover, we observe that these differences are more pronounced in the COVID-19 and COP26 datasets—the topics more closely tied to scientific discourse. Interestingly, this may suggest a higher linguistic differences in scientific topics compared to those of a more socio-political nature such as the Russian-Ukranian conflict.
Ultimately, these results as a whole highlights how users with higher involvement - whether it results from political, ideological or behavioral and civility elements - generally show a more complex vocabulary. These insights help to deep more into the dynamics of online debates, particularly referring to the most active and influential users. 

Our results shed light on the complex relationship between language and influence on social media. Despite its contributions, our study is subject to some limitations that also point to important avenues for future research. First, our analysis is restricted to Twitter/X, which may not generalize to other online spaces.
Second, the study primarily analyzes English-language posts, potentially excluding linguistic nuances or trends present in other languages or regions.
Future studies could extend this work in several directions. First, analyzing language complexity across other social media platforms and in multiple languages would enhance generalizability. Second, longitudinal analyses could reveal how complexity evolves over time in response to events. Finally, incorporating different user classification methods and a wider range of topics may improve accuracy and scope. 




\section{Acknowledgements}
This work is the output of the Complexity72h workshop, held at the Universidad Carlos III de Madrid in Leganés, Spain, 23-27 June 2025 \url{https://www.complexity72h.com}.
We thank Paolo Benanti and Walter Quattrociocchi for inspiring our analysis and project. 

\bibliographystyle{unsrt}







\clearpage

\newcommand{\beginsupplement}{
    \setcounter{section}{0}
    \renewcommand{\thesection}{S\arabic{section}}
    \setcounter{equation}{0}
    \renewcommand{\theequation}{S\arabic{equation}}
    \setcounter{table}{0}
    \renewcommand{\thetable}{S\arabic{table}}
    \setcounter{figure}{0}
    \renewcommand{\thefigure}{S\arabic{figure}}
    \newcounter{SIfig}
    \renewcommand{\theSIfig}{S\arabic{SIfig}}}

\beginsupplement
\section*{\centering Supplementary Information}

\begin{figure*}[ht]
       \centering
    \includegraphics[width=0.5\textwidth]{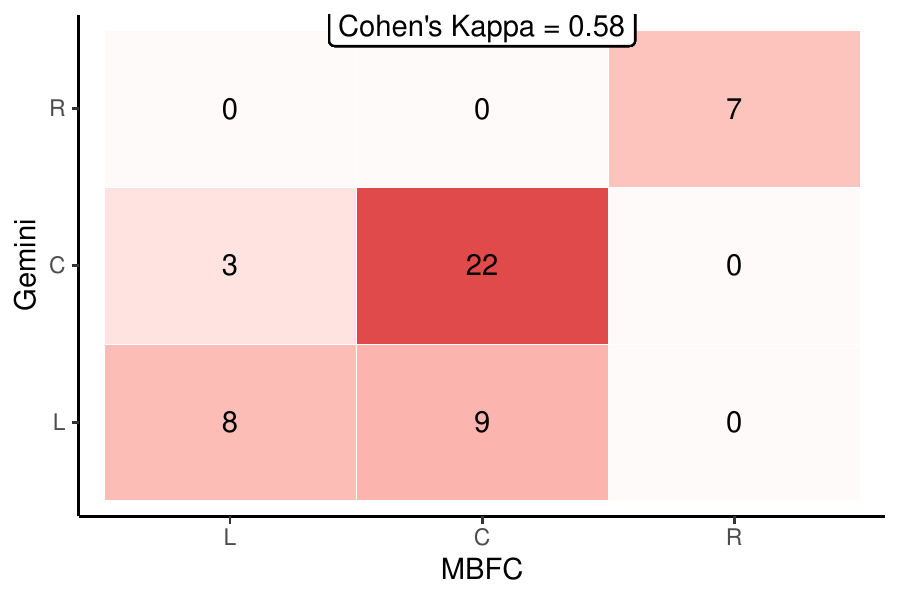}
            \caption{Confusion matrix showing the agreement between the political leaning of user profiles labeled by Gemini and those labeled by MBFC. MBFC ratings were collapsed into broader categories: ``L'' (Left), combining ``left'', ``extreme\_left'', and ``pro\_science''; ``C'' (Center), combining ``center'', ``left\_center'', and ``right\_center''; and ``R'' (Right), combining ``right'' and ``extreme\_right''.}
        \label{fig:fig_confusion_matrix_1_SI}
\end{figure*}

\begin{figure*}[ht]
       \centering
  \includegraphics[width=0.6\textwidth]{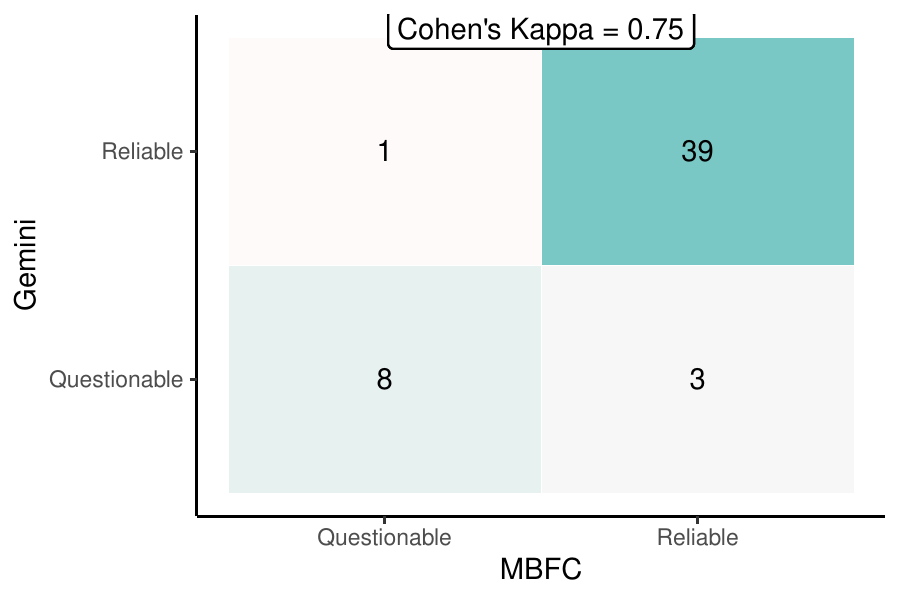}
        \caption{Confusion matrix showing the agreement between the reliability of user profiles labeled by Gemini and those labeled by MBFC.}
        \label{fig:fig_confusion_matrix_2_SI}
\end{figure*}

\begin{table}[ht]
\caption{Results of the Kruskal-Wallis test for the complexity metrics employed in this analysis: \textit{Yule's $K$}, \textit{gzip}, and \textit{Flesch Index}. For each of them, we compare their distributions according to the following features: Account Type, Political Leaning, Reliability, Offensiveness, and Sentiment.}
\label{tab:tab_test}
\centering
\begin{tabular}{cllll}
  \hline
\textbf{Complexity Metric} & \textbf{Feature} & \textbf{Dataset} & \textbf{p-value} \\ 
  \hline
K & Account Type & COP26 & 0.0002 \\ 
  K & Account Type & Ukraine & 0.0018 \\ 
  K & Account Type & COVID-19 & \textless 0.0001 \\ 
  K & Political Leaning & COP26 & 0.4948 \\ 
  K & Political Leaning & Ukraine & 0.5725 \\ 
  K & Political Leaning & COVID-19 & \textless 0.0001 \\ 
  K & Reliability & COP26 & 0.743 \\ 
  K & Reliability & Ukraine & 0.7568 \\ 
  K & Reliability & COVID-19 & \textless 0.0001 \\ 
  K & Offensiveness & COP26 & 0.0044 \\ 
  K & Offensiveness & Ukraine & \textless 0.0001 \\ 
  K & Offensiveness & COVID-19 & \textless 0.0001 \\ 
  K & Sentiment & COP26 & 0.0006 \\ 
  K & Sentiment & Ukraine & \textless 0.0001 \\ 
  K & Sentiment & COVID-19 & \textless 0.0001 \\ 
  GZIP & Account Type & COP26 & \textless 0.0001 \\ 
  GZIP & Account Type & Ukraine & \textless 0.0001 \\ 
  GZIP & Account Type & COVID-19 & \textless 0.0001 \\ 
  GZIP & Political Leaning & COP26 & 0.0007 \\ 
  GZIP & Political Leaning & Ukraine & \textless 0.0001 \\ 
  GZIP & Political Leaning & COVID-19 & \textless 0.0001 \\ 
  GZIP & Reliability & COP26 & 0.0001 \\ 
  GZIP & Reliability & Ukraine & \textless 0.0001 \\ 
  GZIP & Reliability & COVID-19 & \textless 0.0001 \\ 
  GZIP & Offensiveness & COP26 & \textless 0.0001 \\ 
  GZIP & Offensiveness & Ukraine & \textless 0.0001 \\ 
  GZIP & Offensiveness & COVID-19 & \textless 0.0001 \\ 
  GZIP & Sentiment & COP26 & \textless 0.0001 \\ 
  GZIP & Sentiment & Ukraine & \textless 0.0001 \\ 
  GZIP & Sentiment & COVID-19 & \textless 0.0001 \\ 
  Flesch & Account Type & COP26 & \textless 0.0001 \\ 
  Flesch & Account Type & Ukraine & \textless 0.0001 \\ 
  Flesch & Account Type & COVID-19 & \textless 0.0001 \\ 
  Flesch & Political Leaning & COP26 & \textless 0.0001 \\ 
  Flesch & Political Leaning & Ukraine & \textless 0.0001 \\ 
  Flesch & Political Leaning & COVID-19 & \textless 0.0001 \\ 
  Flesch & Reliability & COP26 & 0.0226 \\ 
  Flesch & Reliability & Ukraine & \textless 0.0001 \\ 
  Flesch & Reliability & COVID-19 & \textless 0.0001 \\ 
  Flesch & Offensiveness & COP26 & \textless 0.0001 \\ 
  Flesch & Offensiveness & Ukraine & \textless 0.0001 \\ 
  Flesch & Offensiveness & COVID-19 & \textless 0.0001 \\ 
  Flesch & Sentiment & COP26 & \textless 0.0001 \\ 
  Flesch & Sentiment & Ukraine & \textless 0.0001 \\ 
  Flesch & Sentiment & COVID-19 & \textless 0.0001 \\ 
   \hline
\end{tabular}

\end{table}

\begin{figure*}[ht]
       \centering
    \includegraphics[width=0.5\textwidth]{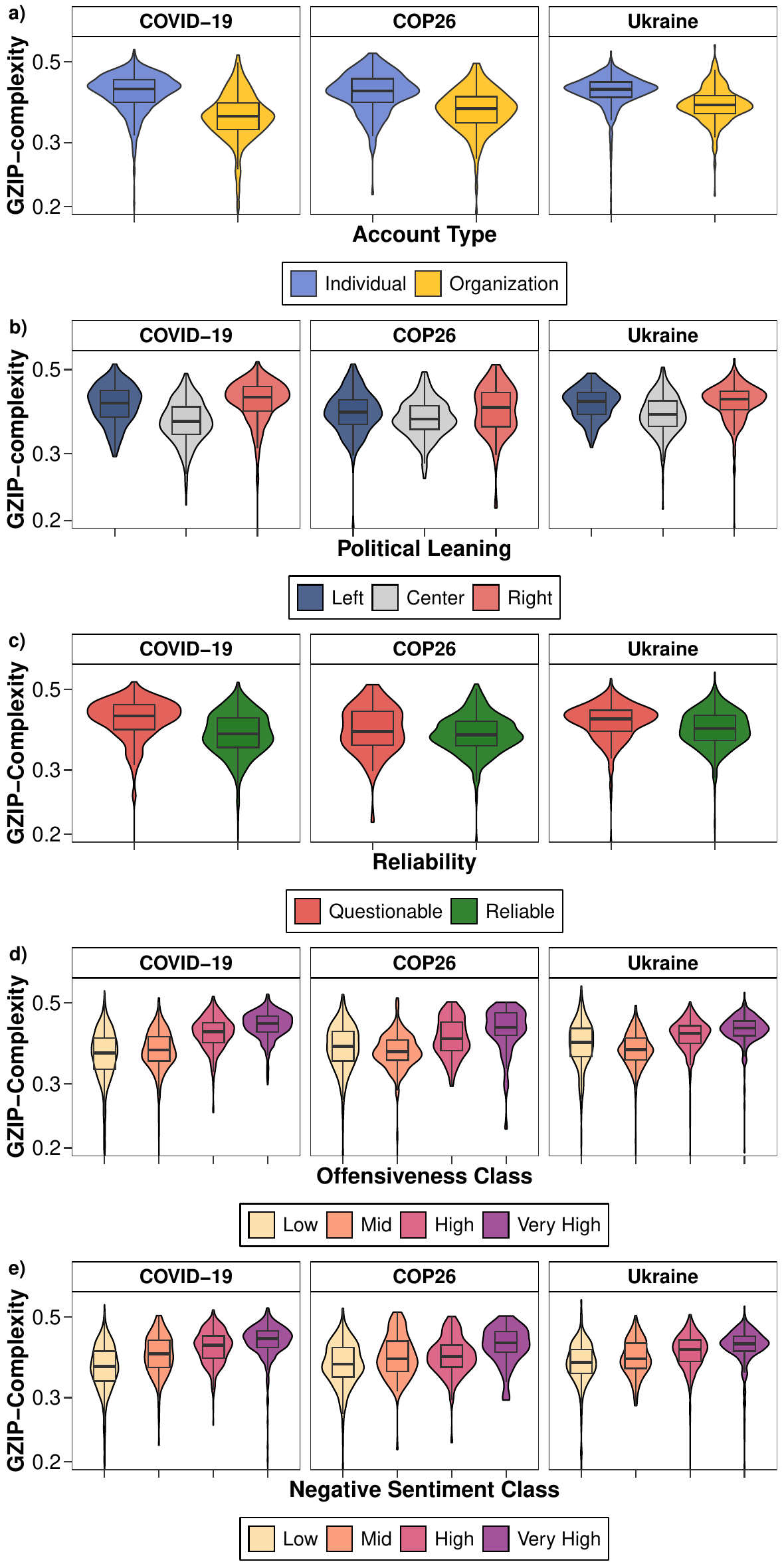}
        \caption{GZIP-complexity distributions according to a) the account type of the influencers (Individual or Organization), b) the political leaning of the influencer (Left, Center, Right), c) the reliability of the influencer (Questionable or Reliable), (d) user offensiveness class and (e) negative sentiment class.}
        \label{fig:fig_gzip_SI}
\end{figure*}

\begin{figure*}[ht]
       \centering
    \includegraphics[width=0.5\textwidth]{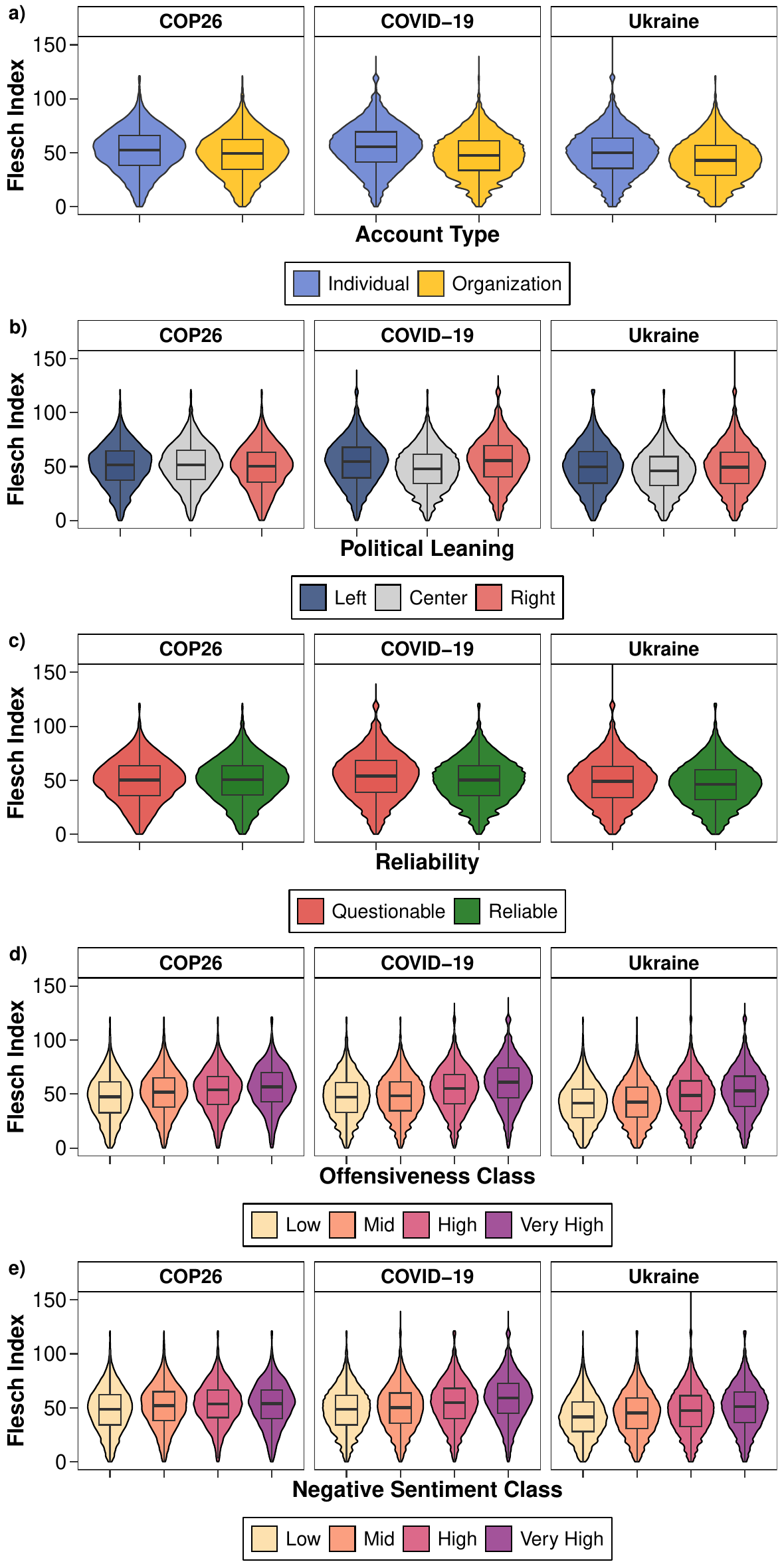}
        \caption{Flesch Index complexity distributions according to a) the account type of the influencers (Individual or Organization), b) the political leaning of the influencer (Left, Center, Right), c) the reliability of the influencer (Questionable or Reliable), (d) user offensiveness class and (e) negative sentiment class.}
        \label{fig:fig_flesch_SI}
\end{figure*}

\clearpage

\begin{table}[ht]
\caption{Information on Bipartite Networks between Influencers and Types for COVID-19, COP26, and Ukraine}
\centering
\begin{tabular}{lrrr}
\hline
\textbf{Metric}           & \textbf{COVID-19} & \textbf{COP26} & \textbf{Ukraine} \\
\hline
Nodes               & 71,953              & 26,918          & 92,030            \\
Nodes (Influencers)             & 1,577               & 561            & 1,166             \\
Nodes (Types)             & 70,376              & 26,357          & 90,864            \\
Edges               & 1,900,389            & 463,720         & 2,347,020          \\
Density                   & 0.00073             & 0.0013           & 0.00055             \\
Average Degree            & 52.82              & 34.45          & 51.01            \\
\hline
\end{tabular}
\vspace{0.1cm}
\label{tab:bipartite_metrics}
\end{table}

\begin{table}[ht]
\caption{Information on Validated Projected Influencer Networks for COVID-19, COP26, and Ukraine}
\centering
\begin{tabular}{lrrr}
\hline
\textbf{Metric}               & \textbf{COVID-19}         & \textbf{COP26}            & \textbf{Ukraine}             \\
\hline
Nodes                   & 1,542                     & 511                       & 1,084                        \\
Edges                   & 137,458                   & 7,748                     & 179,423                      \\
Density                       & 0.12                      & 0.059                      & 0.31                         \\
Average Degree                & 178.29                    & 30.32                     & 331.04                       \\
Average Modularity  & 0.3342 \(\pm\) 0.0003 & 0.3749 \(\pm\) 0.0003    & 0.17149 \(\pm\) 0.00009      \\
\hline
\end{tabular}
\vspace{0.1cm}
\label{tab:network_metrics}
\end{table}

\label{subsec:network}

\end{document}